\documentclass[lettersize,journal]{IEEEtran}
\usepackage{amsmath,amsfonts}
\usepackage{algorithmic}
\usepackage{algorithm}
\usepackage{array}
\usepackage[caption=false,font=normalsize,labelfont=sf,textfont=sf]{subfig}
\usepackage{textcomp}
\usepackage{stfloats}
\usepackage{url}
\usepackage{verbatim}
\usepackage{graphicx}
\usepackage{cite}
\usepackage{multirow}
\begin{document}

\title{Effective Transfer of Pretrained Large Visual Model for Fabric Defect Segmentation via Specific Knowledge Injection}

\author{Zhewei Chen, Wai Keung Wong, Zuofeng Zhong, Jinpiao Liao, Ying Qu
\thanks{Z. Chen is with the School of Fashion and Textiles, The Hong Kong Polytechnic University, Hong Kong, (e-mail: zhewei.chen@connect.polyu.hk).}
\thanks{W.K. Wong is with the School of Fashion and Textiles, The Hong Kong Polytechnic University, Hong Kong, and also with the Laboratory for Artifcial Intelligence in Design, Hong Kong, (e-mail: calvin.wong@polyu.edu.hk)}
\thanks{Corresponding author: W.K. Wong (email: calvin.wong@polyu.edu.hk)}
\thanks{Z. Zhong is with the Laboratory for Artifcial Intelligence in Design, Hong Kong, (e-mail: zhongzuofeng@aidlab.hk)}
\thanks{J. Liao and Y. Qu are with the School of Fashion and Textiles, The Hong Kong Polytechnic University, Hong Kong, (e-mail: jinpiao.liao@connect.polyu.hk; y-ing.qu@connect.polyu.hk).}
}



\maketitle

\begin{abstract}
Fabric defect segmentation is integral to textile quality control. Despite this, the scarcity of high-quality annotated data and the diversity of fabric defects present significant challenges to the application of deep learning in this field. These factors limit the generalization and segmentation performance of existing models, impeding their ability to handle the complexity of diverse fabric types and defects.
To overcome these obstacles, this study introduces an innovative method to infuse specialized knowledge of fabric defects into the Segment Anything Model (SAM), a large-scale visual model. By introducing and training a unique set of fabric defect-related parameters, this approach seamlessly integrates domain-specific knowledge into SAM without the need for extensive modifications to the pre-existing model parameters. The revamped SAM model leverages generalized image understanding learned from large-scale natural image datasets while incorporating fabric defect-specific knowledge, ensuring its proficiency in fabric defect segmentation tasks.
The experimental results reveal a significant improvement in the model's segmentation performance, attributable to this novel amalgamation of generic and fabric-specific knowledge. When benchmarking against popular existing segmentation models across three datasets, our proposed model demonstrates a substantial leap in performance. Its impressive results in cross-dataset comparisons and few-shot learning experiments further demonstrate its potential for practical applications in textile quality control.
\end{abstract}


\section{Introduction}
\IEEEPARstart{F}{abric} defect segmentation plays a crucial role in the quality control of textiles \cite{boluki2021inspection}. As the core product of the textile industry, the quality of fabrics directly affects the overall quality of the final products. Fabric defects, such as holes, stains, and broken yarns, are the main factors affecting fabric quality. Therefore, effectively and accurately identifying and segmenting defects in fabrics is of great significance for ensuring the quality of textiles, improving production efficiency, and reducing production costs \cite{li2021fabric}. Especially in modern textile industry, with the expansion of production scale and the improvement of efficiency, relying on manual identification and segmentation of fabric defects can no longer meet production needs \cite{zhao2020real}. Thus, researching and developing efficient and accurate fabric defect segmentation technologies have become urgent problems to be solved in the field of the textile industry \cite{ngan2011automated}.

\begin{figure*}[!t]
    \centering
    \includegraphics[width=\textwidth]{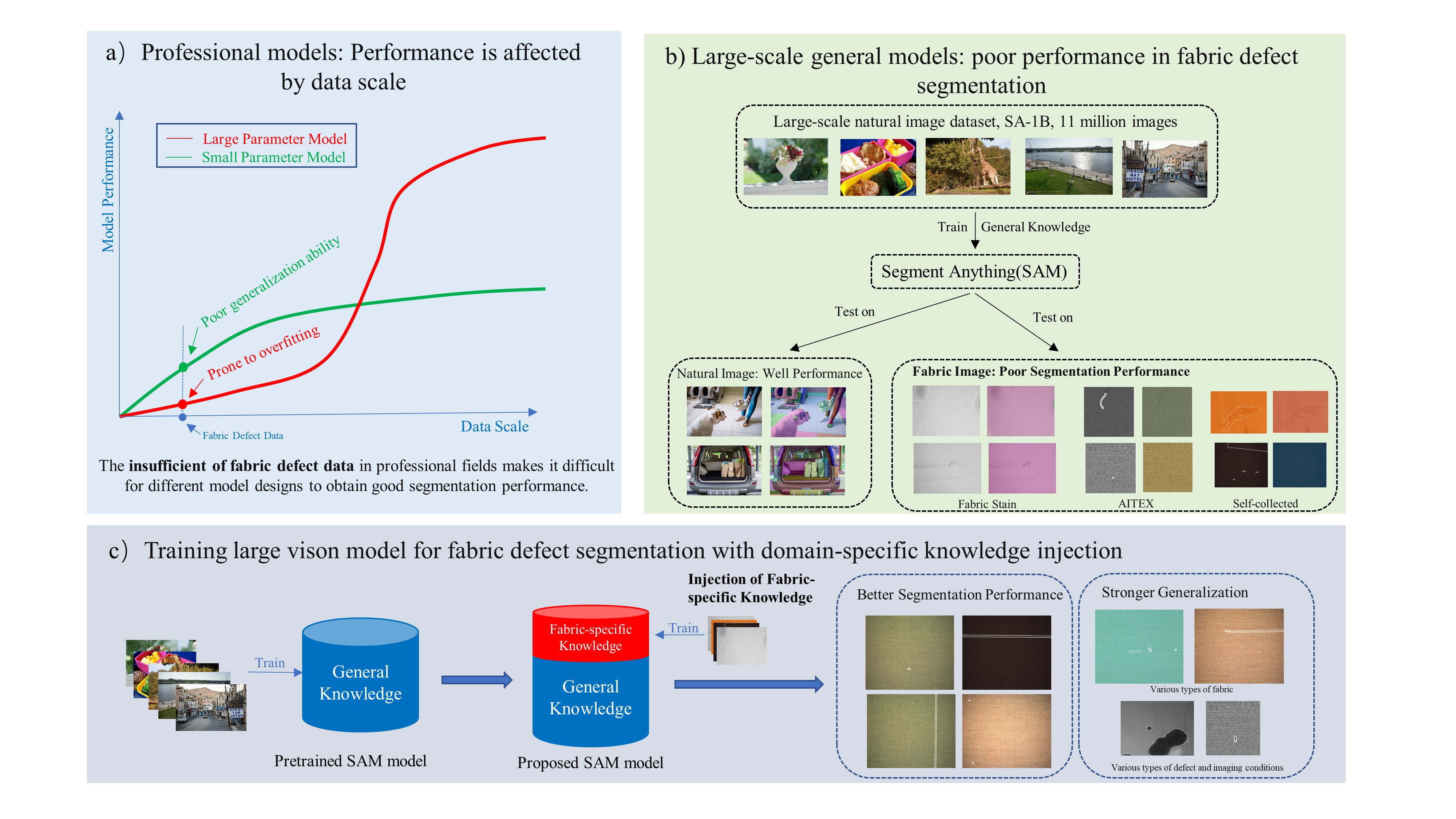}
    \caption{The motivation for this study. a) The scarcity of datasets in the fabric defect domain restricts the performance of specialized models trained solely on fabric defect data. b) Large-scale vision models, trained on vast natural image datasets, lack specialized knowledge of fabric defects, resulting in subpar performance in fabric defect segmentation tasks. c) This paper proposes the injection of specialized fabric defect knowledge into large-scale vision models to enhance their performance and generalization abilities in fabric defect segmentation tasks.}
\label{fig:motivation}
\end{figure*}

Despite the noteworthy successes of deep learning methods in various fabric defect datasets \cite{zhu2020modified,arora2022detection,peng2020automatic,li2016deformable,chen2015mxnet,tian2019autoencoder}, research in fabric defect segmentation still encounters significant hurdles. Apparently, the scarcity of high-quality annotated data, compounded by the wide range of fabric defects and types \cite{9312748}, limits the application of deep learning in this field. Given the extensive diversity of fabric defects and types, creating a fully annotated dataset encompassing all possible combinations is practically unattainable \cite{kahraman2023deep}. This insufficient data thereby hinders the progress of models designed for fabric segmentation tasks. 
As depicted in Figure~\ref{fig:motivation}(a), model with a large number of parameters can capture more complex image features, but they are prone to overfitting in the face of limited data \cite{ying2019overview}, resulting in a decrease in segmentation performance in practical applications. Although models with a small number of parameters can achieve good results on small-scale dataset, often have unsatisfactory generalization ability \cite{canatar2021spectral}, especially when encountering diverse fabric types and defect categories.

In light of these challenges, the recent advent of the Segment Anything (SAM) \cite{kirillov2023segment} presents a potential solution. SAM, a large vision foundation model, leverages training on billions of natural images to accumulate a rich reservoir of generalized knowledge \cite{han2021pre}. This wide-ranging understanding enhances its ability to decode complex image structures and patterns, boosting its generalization capacity. Excelling across diverse image tasks, SAM has earned significant acclaim and wide application \cite{jing2023segment,deng2023segment,ji2023segment,yu2023inpaint}. 
As illustrated in Figure~\ref{fig:motivation}(b), despite the abundant general knowledge inherent within the Segment Anything Model (SAM), its direct application to fabric defect segmentation frequently results in suboptimal performance. This deficiency is principally attributed to the model's insufficiency in specialized knowledge of fabric defects. The general knowledge gained from training on diverse natural images does not effectively translate into identifying and segmenting fabric defects. 
Therefore, the main challenge of transferring the SAM general model to fabric defect segmentation task is how to augment it with fabric defect-specific knowledge while preserving its extensive general knowledge.

To address the aforementioned challenges, a specialized approach is proposed to augment the Segment Anything Model (SAM) with domain-specific knowledge pertaining to fabric defects. In this approach, a novel set of parameters associated with fabric defects is introduced into SAM, utilizing a distinctive training strategy. This strategy, utilizing specific constraints during the training phase, steers these newly introduced parameters towards learning the unique characteristics of fabric defects. Concurrently, these parameters exert subtle influence over the original parameters of SAM, directing them towards an understanding of fabric defects, while conserving their foundational comprehension of visual knowledge. The proposed SAM thus functions as a hybrid model, amalgamating the merits of a large-scale visual model and a task-specific small model. It leverages SAM's extensive generic visual understanding sourced from a variety of natural images, and supplements it with the precise, fabric defect-specific knowledge delivered by the additional parameters. The proposed approach showcases its efficacy in scenarios with limited training data, highlighting the model's robust segmentation performance and generalization capability in the face of data constraints.The contributions of this study are as follows:
\begin{enumerate}
    \item We introduce a novel method that incorporates specialized knowledge of fabric defects into the SAM, maintaining its universal visual comprehension while infusing domain-specific insights. Proposed method design capitalizes on the general knowledge base of large visual model to mitigate issues arising from defect data scarcity.
    
    \item The proposed model has been extensively benchmarked against popular segmentation models across three fabric defect datasets. Supported by a comprehensive set of experimental results, the proposed model has demonstrated outstanding defect segmentation capabilities and impressive generalization ability, thus establishing a significant advancement in fabric quality control.
    
    \item To the best of our knowledge, we are the first to successfully employ a large-scale visual model to the field of fabric defect segmentation. By fully considering the characteristics of the domain, such as the rich texture of textiles, the diversity of defect types, and the scarcity of samples, our method can achieve more accurate and effective detection of fabric defects. This not only enhances the precision and efficiency of fabric defect detection but also provides new insights for research in this field.

\end{enumerate}

This paper is structured as follows: Section 2 reviews related literature. Section 3 elucidates the proposed approach to infuse specialized knowledge into the SAM model. Section 4 describes the experimental design. Sections 5, 6, and 7 present the experimental results, discuss their implications, and summarize the study, respectively.

\section{RELATED WORKS}
\subsection*{A. Fabric Defect Segmentation}
Deep learning has significantly surpassed traditional methods in the task of fabric defect segmentation, with most current methods being based on this approach. These studies can be categorized into two types based on the availability of pixel-wise labeled data. The first category involves supervised learning methods that require pixel-wise labeled data for model training. The second category includes methods that either do not use pixel-wise labeled data or only employ defect-free samples for training. This can be considered as a type of weakly supervised learning method.

In the first type of study, due to the use of pixel-wise training data, many studies have achieved remarkable segmentation performance on corresponding datasets. For example, Cheng et al. \cite{cheng2023fabric} proposed a fabric defect detection method based on Separate Convolutional UNet (SCUNet) and achieved 98.01\% accuracy on AITEX dataset. Huang et al. \cite{huang2021fabric} proposed a highly efficient convolutional neural network for fabric defect detection, which achieved high-accuracy defect localization with as few as 50 defect samples and real-time detection speeds, outperforming eight state-of-the-art methods in terms of accuracy and robustness. Liu et al. \cite{liu2019fabric} proposed a deep saliency model with an attention mechanism for fabric defect detection, successfully enhancing detection performance and surpassing state-of-the-art methods in defect localization without significantly increasing computational demands. Similar studies based on deep learning and supervised learning for fabric defect detection include \cite{jing2022mobile,kopaczka2019detection,sun2019fast,li2016deformable,rong2021fabric,ouyang2019fabric,shao2022pixel}. However, supervised learning heavily relies on large-scale and high-quality labeled data. Collecting and annotating fabric defect images can be challenging, making it impractical to rely solely on supervised learning for fabric defect segmentation.

Some researchers have tried to resolve the issue of pixel-wise annotation acquisition by utilizing more defect-free image information and other types of annotation information for defect segmentation. For instance, Koulali et al. \cite{koulali2021unsupervised} proposed a method that only need one defect-free sample for training and demonstrated competitive performance on the Patterned Fabrics benchmark dataset. Liu et al. \cite{liu2019multistage} proposed a GAN-based framework for fabric defect detection that customizes a deep semantic segmentation network for various defect types and utilizes a multistage GAN to generate reasonable defects on new defect-free samples. Their approach enables continuous dataset updates and contributes to the fine-tuning of the network to enhance defect detection performance under varying conditions. However, these methods often struggle with lower accuracy in defect segmentation due to the difficulty in obtaining precise locations and shapes of defects compared to fully supervised methods. Model training can also be challenging, as weak supervision relies on limited pixel-level annotation data, requiring more iterations and complex optimization algorithms. Performance across different datasets can be unstable, sometimes matching or surpassing supervised methods, while underperforming in other cases \cite{niu2022defect,li2022fabric}.

\subsection*{B. Large-Scale Models}
Large-scale foundational models, characterized by their massive parameter scale and extensive training data, have been receiving widespread attention. The central idea behind these models is to absorb and learn rich feature representations and patterns through pre-training on large-scale datasets, serving as a basis for their deep understanding and handling of complex problems. Such pre-training allows these models to amass a wealth of universal knowledge, which includes a variety of data patterns and deep structures. Therefore, these models can not only be used to solve the tasks they faced during training but can also be fine-tuned for downstream tasks to address specific problems in a given domain. In the field of natural language processing (NLP), renowned pre-trained foundational models includes BERT \cite{devlin2018bert}, GPT-3 \cite{brown2020language}, and GPT-4 \cite{openai2023gpt4}. In the realm of computer vision (CV), models like ResNet \cite{he2016deep}, CLIP \cite{radford2021learning}, and StyleGAN \cite{karras2019style} conduct pre-training on large-scale image datasets, absorbing and learning rich image features, and can subsequently be employed to tackle various image understanding and generation tasks. Through learning on a vast array of image data, these models can effectively extract and understand the deep features of images, thus exhibiting robust performance across numerous vision tasks.

As a new large-scale vision foundational model, SAM, developed by the Meta AI research team, employed a prompt-able segmentation system capable of "cutting out" any object in any image. This model was designed and trained to be promptable. In other words, it can perform zero-shot transfer – that is, it can handle new image distributions and tasks it has never seen before. The core of SAM is a powerful image encoder and a prompt encoder, with the former responsible for computing image embeddings and the latter for embedding prompts. These two information sources are then combined in a lightweight mask decoder to predict segmentation masks, thereby achieving image segmentation tasks.

\subsection*{C. Applications build upon Segment Anything Model (SAM)}
 SAM has been applied in various domains, including medical image segmentation, natural scene segmentation, and video object segmentation. For instance, Deng et al. \cite{deng2023segment} evaluated the SAM for digital pathology tasks and demonstrates that SAM performs well for large connected objects but struggles with dense instance objects. The authors discussed the limitations of SAM in the context of pathological images and proposed future directions for its application. Similarly, Tang et al. \cite{tang2023can} investigated the applicability of the SAM for the task of camouflaged object detection (COD). They evaluated SAM on a COD benchmark using two metrics and compared its performance against 22 state-of-the-art COD methods. The results revealed that SAM exhibits limited performance in the COD task, prompting further research to enhance its generalization to specific scenes. Additionally, in studies \cite{ma2023segment,wu2023medical,zhang2023personalize}, the authors attempted to build foundation model for medical images based on SAM and demonstrated its excellent performance on downstream medical tasks. Similar studies \cite{cen2023sad,mo2023av,zhou2023dsec,larsen2023np} also highlight that while SAM demonstrates impressive performance in the domain of natural images, its segmentation effectiveness diminishes when applied to specific fields such as RGBD images, electron microscopy images and remote sensing images. 

\subsection*{}
To further emphasize the significance of this research, this study utilizes the presence of the rich and generic knowledge in pretrained SAM to compensate for the limited segmentation performance and generalization caused by the insufficient training data in fabric defect segmentation. By integrating the general knowledge from the SAM, the generalization performance of the fabric defect segmentation model is significantly enhanced. Moreover, the learning of specialized fabric defect knowledge ensures the model's ability to accurately recognize fabric defects. To the best of our knowledge, this research is the first to study the application of large-scale pretrained models to the task of fabric defect segmentation.

\section{METHODS}

\begin{figure*}[!t]
    \centering
    \includegraphics[width=\textwidth]{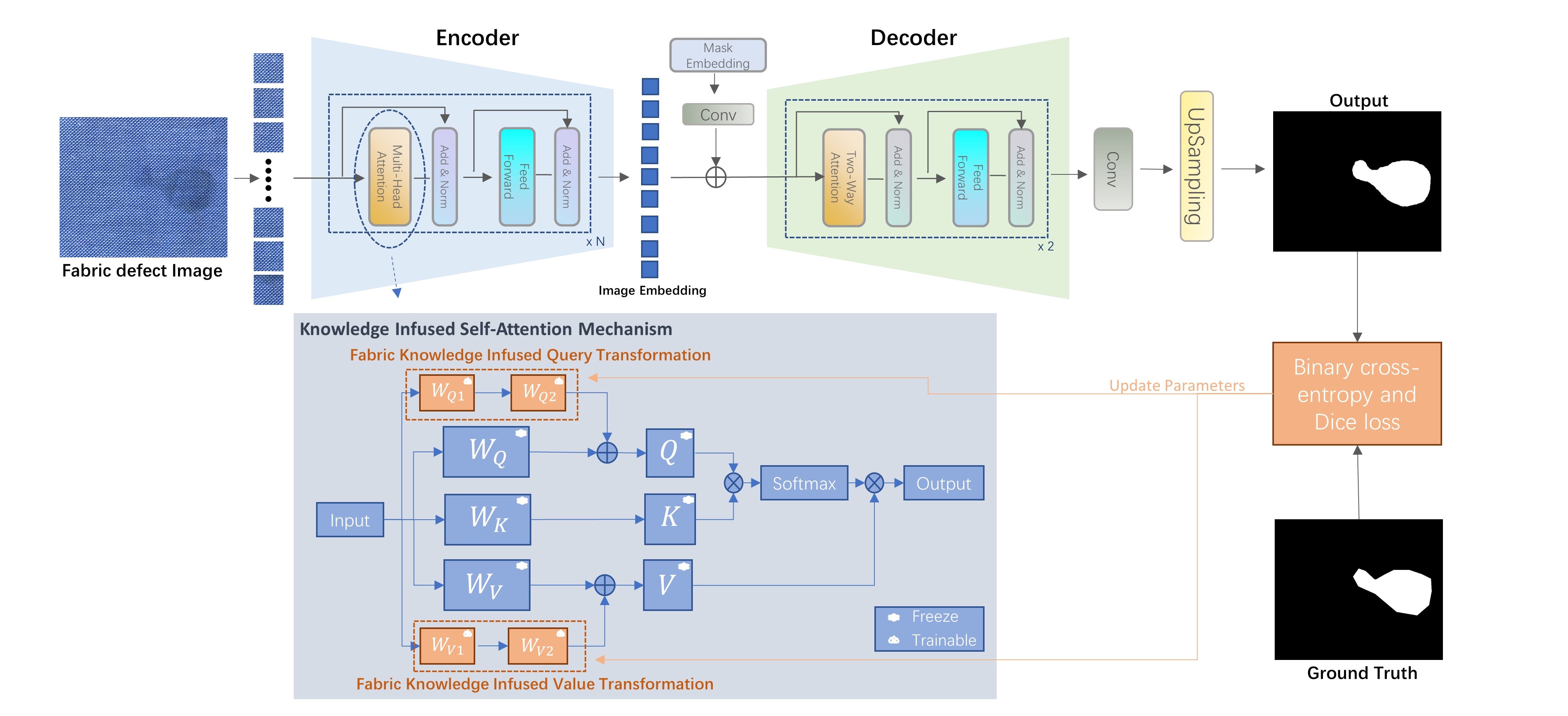}
    \caption{The figure illustrates the architecture of the entire proposed model, with specific modifications made to the self-attention mechanism within the encoder and decoder's transformer modules. Additional matrices, parallel to Wq and Wk, representing the injection of specialized knowledge, have been incorporated into the model. During the training process, only the parameters within these specialized knowledge matrices in the encoder are trainable while other parameters remain frozen. In contrast, all parameters within the decoder are updated.
}
    \label{fig:method}
\end{figure*}

Figure \ref{fig:method} presents the overall architecture of the method proposed in this paper, which is built upon the pretrained Segment Anything (SAM) model. Additional trainable parameters are introduced to the self-attention mechanism in the transformer module of the SAM model, aiming to inject specialized knowledge related to fabric defects. In the following sections, the structure of the SAM model, which serves as a large-scale model, will be discussed. Subsequently, the process of introducing the fabric defect knowledge-related embeddings, as introduced in this paper, will be elaborated. Finally, the optimization process for the entire model will be outlined.

\subsection*{A. The Segment Anything Model (SAM)}

The SAM model is composed of three core components: an Image Encoder, a Prompt Encoder, and a Mask Decoder.

The Image Encoder adopts the classical Vision Transformer architecture \cite{dosovitskiy2020image}. The input image is encoded into image embeddings at a scale of 16x16. During the encoding process, the model introduces two types of attention mechanisms: local and global attention. In local attention, each patch interacts only with other patches within a set window size. Global attention, on the other hand, allows each patch to interact with all other patches. Within the SAM model, the Image Encoder module possesses the largest quantity of parameters. For different versions of SAM (e.g., SAM-B, SAM-L, SAM-H), the model is differentiated by adjusting the embedding dimension and encoder depth; the larger the model scale, the higher the embedding dimension, and the greater the encoder depth.

The role of the Prompt Encoder is to encode image prompts (such as points, boxes, or masks) into a format that can be processed by the decoder. In order to retain the original spatial information, this encoder also adds position encoding. In our experiments, aside from the default dense mask prompts, we did not use other types of prompts. This means that the models in this study do not support interactive segmentation like SAM.

The Mask Decoder is a specialized and modified Transformer encoder structure, which includes a unique dynamic mask generation submodule. In the SAM model, each encoder block utilizes a bidirectional cross-attention mechanism, from prompts to image embeddings and from image embeddings to prompts, so as to effectively learn the interaction between prompts and image embeddings within the model. After going through two encoder blocks, the SAM model upsamples the image embeddings to increase the resolution of the image embeddings. Then, through a Multi-Layer Perceptron (MLP), the model maps the output tokens to a dynamic linear classifier, which is responsible for predicting the mask of the target image. It should be noted that the generated mask is only 1/4 the size of the original image. Therefore, in practical application, we need to upsample it to the same size as the original image.

\subsection*{B. Fabric Defect Knowledge-Related Parameters}

In order to preserve the general knowledge learned by the original SAM model and prevent catastrophic forgetting during the process of learning fabric defect-specific knowledge, additional fabric defect knowledge-related parameters are introduced while retaining the original SAM model architecture. In this study, the integration of general knowledge and fabric defect-specific knowledge is achieved by modifying the self-attention mechanism in the transformer module.

In Transformer structure, the Q, K, and V weight matrices carry the general semantic information of the large-scale model. They play a crucial role in the Multi-Head Attention mechanism, where they compute the matching degree between queries (Q) and keys (K) to determine attention weights. These weights are then applied to the value (V) matrix, generating weighted semantic representations that capture long-range dependencies in the input sequence. These weights can be viewed as a set of encoding functions learned by the model from the input sequence, mapping the inputs into a semantic space and encompassing the model's learning and understanding of general semantic information across different tasks and domains. Consequently, large-scale Transformer models exhibit strong generalization and transfer learning capabilities, delivering outstanding performance across various tasks and domains.

In the original SAM model, the generation process of the Q, K, and V keys in the self-attention mechanism can be expressed using the following equations:
\begin{equation} Q = X \cdot W_Q \end{equation}
\begin{equation} K = X \cdot W_K \end{equation}
\begin{equation} V = X \cdot W_V \end{equation}

Here, \( W_Q \), \( W_K \), and \( W_V \) are weight matrices that are multiplied with the input features X to generate the corresponding Q, K, and V keys. This process leverages the properties of matrix multiplication to map the input features to the space of Q, K, and V keys.

 In this study, the low-rank adaptation technique \cite{hu2021lora} is employed to modify the generation of Q, K, and V keys in the model. This technique involved adding a bypass to the original weight matrices to achieve low-rank approximation. The schematic diagram of knowledge infused self-attention mechanism in Figure~\ref{fig:method} illustrates this process. The generation of Q, K, and V keys can be formulated as the combination of the original weight matrix and additional linear layers \( W_{Q1} \), \( W_{Q2} \), \( W_{V1} \), and \( W_{V2} \) obtained from the bypass. It can be formulated as:
 
\begin{equation} Q = X \cdot W_{Q1} \cdot W_{Q2} + X \cdot W_Q\end{equation}
\begin{equation} K = X \cdot W_K \end{equation}
\begin{equation} V = X \cdot W_{V1} \cdot W_{V2} + X \cdot W_V\end{equation}

In the above formulas, X represents the input features. The output of \( W_{Q1} \) and \( W_{Q2} \) actually forms a new weight matrix that represents the newly learned knowledge specific to the fabric defect detection task. In this context, \( W_{Q1} \) and \( W_{Q2} \) can be seen as components of the weight matrix, which are trained to capture specific features relevant to fabric defect detection. In this setup, the original weight matrix \( W_Q \) retains the general knowledge learned from pre-trained SAM, while \( W_{Q1} \cdot W_{Q2} \) provides task-specific knowledge for fabric defect detection.
In this case, the dimensions of \(W_{Q1}\) and \(W_{Q2}\) can be represented as \(W_{Q1} \in \mathbb{R}^{in \times r}\) and \(W_{Q2} \in \mathbb{R}^{r \times out}\), respectively. Here, \(in\) represents the dimensionality of the input features, \(out\) represents the dimensionality of the output features, and it should match the dimensionality of \(W\). \(R\) denotes the dimensionality of the low-rank adaptation, and an appropriate value of \(r\) can be chosen to reduce the parameter count while maintaining model performance.

After obtaining the new Q, K, and V matrices, the self-attention mechanism remains consistent with the original SAM and can be expressed by the following equation:

\begin{equation}
\text{Attention}(Q, K, V) = \text{softmax}\left(\frac{QK^T}{\sqrt{d_k}}\right) V
\end{equation}

Here, Q, K, and V represent the query matrix, key matrix, and value matrix, respectively. By calculating the similarity between the queries and keys, normalizing the similarity scores using the Softmax function, and applying the resulting attention weights to the value matrix, a weighted semantic representation is generated.

\subsection*{C. Overall Training Strategy}

The fabric defect segmentation problem can be formulated as a pixel-wise binary classification problem. Given an input fabric image \(I\) with a resolution of \(H \times W\), the model generates a segmentation map \(S\) of the same resolution. Each pixel in \(S\) is classified as either defective or non-defective.
Mathematically, the function that our model needs to learn can be described as follows:
\begin{equation}
F: I \in \mathbb{R}^{H \times W \times 3} \rightarrow S \in \{0, 1\}^{H \times W}
\end{equation}
Here, \(F\) is the function implemented by our model, \(3\) is the number of color channels in the input image, and \(\{0, 1\}\) represents the binary label for each pixel, where \(0\) denotes non-defective and \(1\) denotes defective.

During the training process, we modify each self-attention layer in both the encoder and decoder to the structure illustrated in Figure~???\ref{fig:method}. In the encoder, except for the additional layers, all parameters are fixed, and only the parameters of the introduced additional layers are updated. As for the decoder, we update all parameters.

 A composite loss function that combines the Dice loss and Binary Cross-Entropy (BCE) loss is used in training process. This selection effectively balances the challenges of class imbalance and spatial information loss. For a single image, the loss function \(L\) can be represented as:

\begin{equation}
L = \alpha L_{BCE} + (1-\alpha) L_{Dice}
\end{equation}

Here, \(L_{BCE}\) refers to the Binary Cross-Entropy loss, calculated as follows:

\begin{equation}
L_{BCE} = -\frac{1}{H \times W} \sum_{i=1}^{H} \sum_{j=1}^{W} [y_{ij} \log(p_{ij}) + (1-y_{ij}) \log(1-p_{ij})]
\end{equation}

In this equation, \(y_{ij}\) is the true label of the pixel at location \((i, j)\), and \(p_{ij}\) is the predicted probability that the pixel at location \((i, j)\) is defective.

Meanwhile, \(L_{Dice}\) refers to the Dice loss, computed as follows:

\begin{equation}
L_{Dice} = 1 - \frac{2 \sum_{i=1}^{H} \sum_{j=1}^{W} y_{ij} p_{ij} + \epsilon}{\sum_{i=1}^{H} \sum_{j=1}^{W} y_{ij}^2 + \sum_{i=1}^{H} \sum_{j=1}^{W} p_{ij}^2 + \epsilon}
\end{equation}

In this equation, \(\epsilon\) is a small number added to prevent division by zero. \(\alpha\) is a weight parameter controlling the relative importance of the two loss components.

\section{EXPERIMENTS}
\subsection*{A. Experiment Data}
In this study, three fabric defect detection datasets: the Fabric Stain dataset \cite{pathirana2020fabric}, the AITEX \cite{silvestre2019public} dataset, and a self-collected dataset are utilized. The Fabric Stain and AITEX datasets are publicly accessible, while the self-collected dataset was procured from a weaving factory. 
The Fabric Stain dataset originally provides bounding box annotations for defects and the pixel-level annotations is subsequently generated based on these bounding boxes. Both the AITEX and self-collected datasets incorporate pixel-level annotations, precisely corresponding to the dimensions of the original images.
The datasets after the filtration process are as follows: The Fabric Stain dataset comprises 394 defective images and corresponding labels, the AITEX dataset contains 185 defective images and labels, and the self-collected dataset consists of 1515 defective images and labels. Each dataset was then partitioned into distinct subsets: 60\% of the images for training, 20\% for validation, and the remaining 20\% for testing.

\subsection*{B. Evaluation Metrics}
To assess fabric defect segmentation, four widely-used metrics, recall, precision, Dice coefficient (Dice), and Intersection over Union (IoU), are used:

\textbf{Recall} measures the proportion of correctly identified positive samples:
\begin{equation} \text{Recall} = \frac{\text{TP}}{\text{TP} + \text{FN}} \end{equation}
\textbf{Precision} assesses the proportion of correctly identified positive predictions:
\begin{equation} \text{Precision} = \frac{\text{TP}}{\text{TP}+\text{FP}} \end{equation}
\textbf{Dice Coefficient} gauges the overlap between the predicted and actual segmentation:
\begin{equation} Dice = \frac{2 \times TP}{2 \times TP + FP + FN} \end{equation}
\textbf{Intersection over Union (IoU)} evaluates the ratio of the intersection to the union of predicted and actual areas:
\begin{equation} IoU = \frac{TP}{TP + FP + FN} \end{equation}
Where TP is True Positive, FP is False Positive, and FN is False Negative.

\subsection*{C. Training Process}
In all experiments, the inputs to all models are resized to 512x512 using the letterbox method to facilitate cross-validation across different datasets. If the original image size is less than 512x512, no magnification is performed, and padding is added instead. The optimization process of all models, including our proposed SAM, employs the AdamW optimizer with momentum correction to update the model parameters, where the learning rate is set to 1e-3 and the beta values are set to 0.937 and 0.999. The training procedure uses a mini-batch of 8 samples and is executed on a GPU with 24GB of memory. If a single GPU cannot handle a batch of 8 images at once, we use a gradient accumulation approach for updates. A cosine annealing warm restarts learning rate scheduler is also employed in this study. It has a period of 10, a period multiplier of 100, and a minimum learning rate set to one percent of the initial learning rate. The model is trained for 1000 epochs and utilizes an early stopping strategy, saving checkpoints at the epoch where the model achieves the best validation Dice score. All non-SAM models were implemented with a ResNet101 \cite{he2016deep} backbone and initialized with weights pre-trained on the ImageNet dataset \cite{deng2009imagenet}.

During the training process, a series of data augmentation techniques is incorporated to enhance the generalization capabilities of the model. Specifically, the augmentation pipeline includes horizontal and vertical flipping of images, each with a probability of 0.5. Additionally, a random 90-degree rotation is applied with a probability of 0.5. Moreover, random adjustments of brightness and contrast are utilized, each within a range of 0.2 and with a likelihood of 0.5.

\subsection*{D. Cross-Dataset Validation}

In the cross-validation experiment, the model was trained on one dataset and subsequently tested on the other two datasets, rotating through each of the datasets in this manner. For example, when the self-collected dataset was used for training, the entire Fabric Stain and AITEX datasets were used for testing. This procedure was then repeated for the other datasets, providing a robust measure of the model's generalization capability.
During the validation process, the original hyperparameters and training strategies used during the training phase were strictly maintained. This approach was crucial in ensuring that the model's performance on different datasets could be attributed to its generalization ability, rather than overfitting or specific adjustments tailored to a particular dataset.

\subsection*{E. Few-Shot Experiments}

This experiment aims to evaluate the performance of the SAM model in limited-sample learning scenarios. Specifically, the three variants of the SAM model (SAM-B, SAM-L, and SAM-H) were tested in three different scenarios (10 samples, 50 samples, and 100 samples) using our self-collected dataset.

In each scenario, a random selection of 10, 50, or 100 samples was made from the training set of the self-collected dataset for model training. Subsequently, the model was tested on the original test set. The training process in each scenario remained consistent with the process used when utilizing the full image set.

\section{RESULTS}

% \usepackage{multirow}
\begin{table*}[ht]
\centering
\caption{Comparison of Segmentation Performance of Various Methods on Different Datasets}
\resizebox{\linewidth}{!}{%
\renewcommand{\arraystretch}{1.2}
\begin{tabular}{|c|c|cccc|cccc|cccc|}
\hline
\multirow{3}{*}{Train   on} & \multirow{3}{*}{Method} & \multicolumn{12}{c|}{Test on} \\
\cline{3-14}
 &  & \multicolumn{4}{c|}{Stain} & \multicolumn{4}{c|}{AITEX} & \multicolumn{4}{c|}{Self-collected} \\
 \cline{3-14}
 &  & recall & precision & f1 & iou & recall & precision & f1 & iou & recall & precision & f1 & iou \\
  \hline
\multirow{12}{*}{Stain} & Unet & 0.8212 & 0.7328 & 0.7452 & 0.6452 & 0.5386 & 0.1278 & 0.1144 & 0.0812 & 0.4873 & 0.0759 & 0.0641 & 0.045 \\
 & Unet++ & 0.8431 & 0.764 & 0.7748 & 0.6849 & 0.5765 & 0.3908 & 0.1539 & 0.1202 & \textbf{0.5761} & \textbf{0.259} & 0.1833 & 0.128 \\
 & Manet & 0.8496 & 0.7694 & 0.7716 & 0.6783 & 0.5476 & 0.5156 & 0.0765 & 0.0758 & 0.5403 & 0.4087 & 0.0953 & 0.0883 \\
 & Linknet & 0.8318 & 0.7609 & 0.7769 & 0.6743 & 0.5777 & 0.2296 & 0.1922 & 0.1201 & 0.5331 & 0.1811 & 0.1613 & 0.0911 \\
 & FPN & 0.8502 & 0.7606 & 0.7646 & 0.6871 & 0.5896 & 0.3305 & 0.1348 & 0.1192 & 0.5688 & 0.2407 & 0.1927 & 0.1378 \\
 & PSPNet & 0.821 & 0.7247 & 0.7333 & 0.6357 & 0.5775 & 0.2613 & 0.0797 & 0.0753 & 0.4747 & 0.0141 & 0.0088 & 0.0061 \\
 & PAN & 0.8538 & 0.7449 & 0.7263 & 0.651 & 0.5555 & 0.3151 & 0.0906 & 0.0846 & 0.476 & 0.0255 & 0.0247 & 0.015 \\
 & DeepLabV3 & 0.863 & 0.7656 & 0.7598 & 0.6884 & \textbf{0.6011} & 0.2167 & 0.1071 & 0.0909 & 0.5146 & 0.1307 & 0.1289 & 0.0785 \\
 & DeepLabV3+ & 0.8254 & 0.7467 & 0.7628 & 0.6646 & 0.5929 & 0.412 & 0.1064 & 0.1 & 0.5565 & 0.2248 & 0.1295 & 0.0921 \\
 & Proposed SAM-B & 0.8678 & 0.7945 & 0.7978 & 0.7157 & 0.5609 & 0.1864 & 0.1859 & 0.1414 & 0.5418 & 0.1863 & 0.1903 & 0.1401 \\
 & Proposed SAM-L & 0.8911 & 0.8215 & 0.8193 & 0.7536 & 0.5885 & 0.2294 & 0.2201 & 0.1748 & 0.5547 & 0.2191 & 0.2254 & 0.1673 \\
 & Proposed SAM-H & \textbf{0.8992} & \textbf{0.8356} & \textbf{0.8327} & \textbf{0.7673} & 0.5358 & \textbf{0.5225} & \textbf{0.2269} & \textbf{0.1869} & 0.5588 & 0.2322 & \textbf{0.2391} & \textbf{0.1777} \\
 \hline
\multirow{12}{*}{AITEX} & Unet & 0.4659 & 0.0669 & 0.069 & 0.0081 & 0.6625 & 0.4025 & 0.409 & 0.3162 & 0.4954 & 0.1262 & 0.1108 & 0.0381 \\
 & Unet++ & 0.4976 & 0.198 & 0.21 & 0.079 & 0.6978 & 0.494 & 0.4709 & 0.3604 & 0.5007 & 0.2175 & 0.1094 & 0.0507 \\
 & Manet & 0.504 & 0.1958 & 0.2084 & 0.09 & 0.6935 & 0.4695 & 0.4558 & 0.3354 & 0.4762 & 0.0604 & 0.0556 & 0.0211 \\
 & Linknet & 0.4903 & 0.1827 & 0.1939 & 0.0622 & 0.6645 & 0.4159 & 0.4123 & 0.3029 & 0.4867 & 0.13 & 0.1332 & 0.0382 \\
 & FPN & 0.4801 & 0.0885 & 0.0904 & 0.0327 & 0.6804 & 0.4459 & 0.4479 & 0.3466 & 0.498 & 0.167 & 0.1545 & 0.0521 \\
 & PSPNet & 0.4719 & 0.035 & 0.0377 & 0.0197 & 0.6197 & 0.3025 & 0.3069 & 0.2413 & 0.476 & 0.0464 & 0.0486 & 0.0168 \\
 & PAN & 0.4821 & 0.1204 & 0.1155 & 0.0237 & 0.6692 & 0.4591 & 0.4786 & 0.3482 & 0.4745 & 0.0439 & 0.0453 & 0.0149 \\
 & DeepLabV3 & 0.4713 & 0.045 & 0.042 & 0.0138 & 0.5955 & 0.2985 & 0.3088 & 0.2111 & 0.5097 & 0.1544 & 0.1558 & 0.0728 \\
 & DeepLabV3+ & 0.5392 & 0.2236 & 0.2212 & 0.132 & 0.6515 & 0.4244 & 0.4173 & 0.2767 & 0.5393 & 0.1774 & 0.1437 & 0.0712 \\
 & Proposed SAM-B & 0.6501 & 0.5755 & 0.615 & 0.383 & 0.6903 & \textbf{0.5006} & \textbf{0.5208} & 0.3941 & 0.5492 & 0.2953 & 0.3144 & 0.1564 \\
 & Proposed SAM-L & 0.6503 & 0.5871 & 0.6248 & 0.3749 & \textbf{0.7396} & 0.4942 & 0.4698 & \textbf{0.3984} & \textbf{0.6021} & \textbf{0.4452} & \textbf{0.47} & \textbf{0.2722} \\
 & Proposed SAM-H & \textbf{0.6518} & \textbf{0.6062} & \textbf{0.6539} & \textbf{0.4005} & 0.6741 & 0.466 & 0.4839 & 0.3576 & 0.5564 & 0.3965 & 0.4337 & 0.2116 \\
 \hline
\multirow{12}{*}{Self-collected} & Unet & 0.5147 & 0.244 & 0.2625 & 0.1155 & 0.552 & 0.2818 & 0.1502 & 0.1158 & 0.74 & 0.6044 & 0.6145 & 0.4991 \\
 & Unet++ & 0.4737 & 0.1099 & 0.1151 & 0.0256 & 0.6494 & 0.3555 & 0.2806 & 0.2191 & 0.7175 & 0.5231 & 0.5306 & 0.4433 \\
 & Manet & 0.5582 & 0.3265 & 0.345 & 0.1932 & 0.6954 & 0.4781 & 0.2459 & 0.2269 & 0.7599 & 0.6399 & 0.6626 & 0.5383 \\
 & Linknet & 0.5088 & 0.2167 & 0.212 & 0.1111 & 0.5861 & 0.2753 & 0.2156 & 0.1485 & 0.7181 & 0.5802 & 0.6127 & 0.4739 \\
 & FPN & 0.5744 & 0.3044 & 0.3136 & 0.2045 & 0.5888 & 0.2403 & 0.1751 & 0.1345 & 0.7442 & 0.6005 & 0.6262 & 0.5061 \\
 & PSPNet & 0.566 & 0.3228 & 0.3418 & 0.2052 & 0.5014 & 0.0803 & 0.0853 & 0.0528 & 0.7076 & 0.5367 & 0.5646 & 0.444 \\
 & PAN & 0.5882 & 0.3535 & 0.3659 & 0.2315 & 0.5829 & 0.2518 & 0.2142 & 0.1374 & 0.7526 & 0.6373 & 0.6682 & 0.5289 \\
 & DeepLabV3 & 0.4976 & 0.1363 & 0.1427 & 0.0706 & 0.5603 & 0.1833 & 0.1476 & 0.1245 & 0.7505 & 0.6179 & 0.6455 & 0.5222 \\
 & DeepLabV3+ & 0.5289 & 0.2833 & 0.3055 & 0.1432 & 0.6368 & 0.405 & 0.1149 & 0.137 & 0.724 & 0.5667 & 0.593 & 0.4729 \\
 & Proposed SAM-B & 0.7701 & 0.6501 & 0.6485 & 0.5313 & \textbf{0.6907} & \textbf{0.4195} & \textbf{0.3873} & 0.3063 & 0.7738 & 0.6948 & 0.7307 & 0.5727 \\
 & Proposed SAM-L & \textbf{0.8462} & \textbf{0.7627} & \textbf{0.7568} & \textbf{0.6688} & 0.6653 & 0.3776 & 0.3266 & 0.2778 & \textbf{0.7855} & 0.6944 & 0.7257 & \textbf{0.589} \\
 & Proposed SAM-H & 0.8288 & 0.7424 & 0.7447 & 0.646 & 0.6722 & 0.3939 & 0.379 & \textbf{0.3072} & 0.7791 & \textbf{0.703} & \textbf{0.7403} & 0.5859\\
 \hline
 
\end{tabular}
}
\label{table:different_method}
\end{table*}

Table~\ref{table:different_method} presents a comprehensive comparative analysis of the segmentation performance of our proposed models (SAM-B, SAM-L, SAM-H) against several other prominent models in the field, such as Unet \cite{ronneberger2015u}, Unet++ \cite{zhou2018unet++}, Manet \cite{9201310}, Linknet \cite{chaurasia2017linknet}, FPN \cite{kirillov2017unified}, PSPNet \cite{zhao2017pyramid}, PAN \cite{li2018pyramid}, DeepLabV3 \cite{chen2017rethinking}, and DeepLabV3+ \cite{chen2018encoder}. The models were tested on three different fabric defect detection datasets: Stain, AITEX, and a self-collected dataset. In Table~\ref{table:different_method}, the results on the diagonal show the performance of models when trained and tested on the same dataset. In contrast, the other elements reflect the results of cross-dataset testing, demonstrating the model's performance when trained on one dataset and tested on another.

In the experiments that training and testing on the same dataset, the proposed SAM models consistently demonstrate superior performance over other models.
For the Stain Dataset, the proposed SAM-H model significantly outperformes all other models across all performance metrics. It achieves substantial improvements in recall, precision, F1 score, and IoU compared to the next best-performing models. Specifically, the SAM-H model surpasses other models by 3.6\% in recall, 7\% in precision, 5.8\% in F1 score, and 8\% in IoU.
On the AITEX dataset, the proposed SAM models continue to exhibit the highest segmentation performance. Notably, the SAM-L model yieldes the best results in terms of recall and Intersection over Union (IoU), while the SAM-B model achieves the best precision and F1 score. In terms of segmentation metrics, the proposed SAM models outperform the non-SAM models by 4.2\%, 0.6\%, 4.2\%, and 3.8\%, respectively.
For the self-collected dataset, the proposed SAM models again obtain the best segmentation results. In particular, the SAM-H model demonstrates superior performance in terms of precision and F1 score, while the SAM-L model achieved the best recall and IOU. The proposed SAM models exhibit improved segmentation performance compared to the non-SAM models, with enhancements of 2.6\% in recall, 6.3\% in precision, 5.1\% in IoU, and 7.2\% in F1 score.

From the results presented, it is evident that the proposed SAM models considerably outperform the non-SAM segmentation models across all metrics and datasets. This underpins the precision and effectiveness of the SAM approach in fabric defect segmentation. Observing the performance trends, it can be noted that except for the AITEX dataset, where the SAM-B model achieves the best F1 score, the SAM-L model, which has the largest number of parameters, yielded the highest F1 score on all other datasets. 

% \usepackage{multirow}
\begin{table*}[ht]
\centering
\caption{Performance metrics of the proposed SAM-B, SAM-L, and SAM-H models on the Self-collected dataset under few-shot learning conditions with 10, 50, 100 training samples, and full training set scenario.}
\resizebox{\linewidth}{!}{%
\renewcommand{\arraystretch}{1.2}
\begin{tabular}{|c|c|c|ccccc|}
\hline
Dataset & \multicolumn{1}{c|}{Training Samples} & Model & Accuracy & Recall & Precision & F1 & Iou \\
\cline{1-8}
\multirow{12}{*}{Self-collected} & \multirow{3}{*}{10-shots} & Proposed SAM-B & 0.9852 & 0.6396 & 0.3995 & 0.4131 & 0.2876 \\
 &  & Proposed SAM-L & 0.9838 & 0.6923 & 0.5463 & 0.5755 & 0.4109 \\
 &  & Proposed SAM-H & 0.9343 & 0.6312 & 0.3559 & 0.3348 & 0.2618 \\
 \cline{2-8}
 & \multirow{3}{*}{50-shots} & Proposed SAM-B & 0.9903 & 0.7498 & 0.5733 & 0.5807 & 0.481 \\
 &  & Proposed SAM-L & 0.9882 & 0.772 & 0.6543 & 0.675 & 0.5445 \\
 &  & Proposed SAM-H & 0.9902 & 0.7578 & 0.6227 & 0.6443 & 0.5198 \\
 \cline{2-8}
 & \multirow{3}{*}{100-shots} & Proposed SAM-B & 0.9917 & 0.7329 & 0.6079 & 0.6405 & 0.49 \\
 &  & Proposed SAM-L & 0.9914 & 0.7559 & 0.6507 & 0.6845 & 0.5343 \\
 &  & Proposed SAM-H & 0.9915 & 0.7381 & 0.6006 & 0.627 & 0.4929 \\
 \cline{2-8}
 & \multirow{3}{*}{Full training} & Proposed SAM-B & 0.9948 & 0.7738 & 0.6948 & 0.7307 & 0.5727 \\
 &  & Proposed SAM-L & 0.9949 & 0.7855 & 0.6944 & 0.7257 & 0.589 \\
 &  & Proposed SAM-H & 0.9948 & 0.7791 & 0.703 & 0.7403 & 0.5859 \\
 \cline{1-8}
\end{tabular}
}
\label{table:few-shot}
\end{table*}

Table ~\ref{table:different_method} presents the results of different models in cross-dataset validation. In the cross-dataset cross-validation experiments, F1 score, as a comprehensive indicator of segmentation effectiveness, is the primary focus. This is because cross-dataset experiments often yield all 0 or all 1 results in predictions (i.e., no predictive capability), making some of the indicators less reliable.

When the models trained on the Stain dataset and test on the AITEX dataset, the SAM-H model achieves the best F1 score, surpassing the strongest non-SAM model, Unet++, by 7.3\%. When testing on the self-collected dataset, SAM-h achieves the best F1 score, surpassing the best non-SAM model, FPN, by 4.6\%. Although the SAM model still achieves the best segmentation results, the overall performance is not satisfactory. This could be due to the fact that the Stain dataset contains only the stain category of defects, leading to unsatisfactory performance on tests involving other types of defects.

When the models trained on the AITEX dataset and test on the Stain dataset, the SAM-h model achieves the best F1 score, surpassing the best non-SAM model by 43.3\%. When testing on the self-collected dataset, the SAM-l model achieves the best F1 score, exceeding the best non-SAM model, DeepLabV3, by 31.4\%. The results showes that the cross-database prediction of non-SAM models is unsuccessful. This is because the AITEX database has a relatively low number of training images (approximately over 100), including only one type of defect, and has an image resolution of only 256x256. Such limited training data quality resulted in poor performance in cross-dataset validation for other non-SAM models, both on the stain and self-collected datasets (F1 score no more than 0.2). In contrast,the proposed SAM model demonstrates stronger segmentation performance on the stain dataset and a certain degree of segmentation capability on the self-collected dataset. These cross-validation results underscore the exceptional generalization ability of our proposed SAM model, capable of accurately segmenting noticeable defects even with limited training data quality. This also showcases the zero-shot capability of our proposed SAM model from another perspective, as the AITEX training data does not include stain-type defects.

Finally, when training on the self-collected dataset and testing on the Stain dataset, the SAM-L model achieves the best F1 score, surpassing the best non-SAM model, PAN, by 39.1\%. When testing on the AITEX dataset, the SAM-b model achieves the best overall performance, exceeding the best non-SAM model, Unet++, by 10.7\%. In particular, on the stain dataset, the SAM model trained on the self-collected dataset achieves a segmentation performance similar to that of non-SAM models trained on the stain dataset.

Table~\ref{table:few-shot} presents the performance indicators of the proposed models (SAM-B, SAM-L, SAM-H) under different few-shot learning conditions (10, 50, and 100 training samples, as well as the full training set of self-collected dataset) on the self-collected dataset. From the experiment results,it is observed that with an increase in the number of training samples from 10 to 50, the F1 scores of the SAM-B, SAM-L, and SAM-H models show significant improvement. Specifically, under the 10-shot scenario, the F1 scores are 0.4131, 0.5755, and 0.3348 for SAM-B, SAM-L, and SAM-H models respectively. However, under the 50-shot scenario, the F1 scores improve to 0.5807, 0.675, and 0.6443 for SAM-B, SAM-L, and SAM-H models respectively. Further increasing the number of training samples to 100 leads to more improvements, albeit not as significant as the previous scenario. In this 100-shot scenario, the F1 scores for SAM-B, SAM-L, and SAM-H are 0.6405, 0.6845, and 0.627 respectively. When the entire dataset (909 samples) is used for training, the performance of all models improve slightly, with F1 scores for SAM-B, SAM-L, and SAM-H being 0.7307, 0.7257, and 0.7403 respectively. Furthermore, upon examining the few-shot results, it is observed that the proposed SAM models under the 50-shot scenario are able to achieve performance levels comparable to non-SAM models trained under full-sample conditions.
 
\section{DISCUSSION}

\begin{figure*}[!t]
    \centering
    \includegraphics[width=\textwidth]{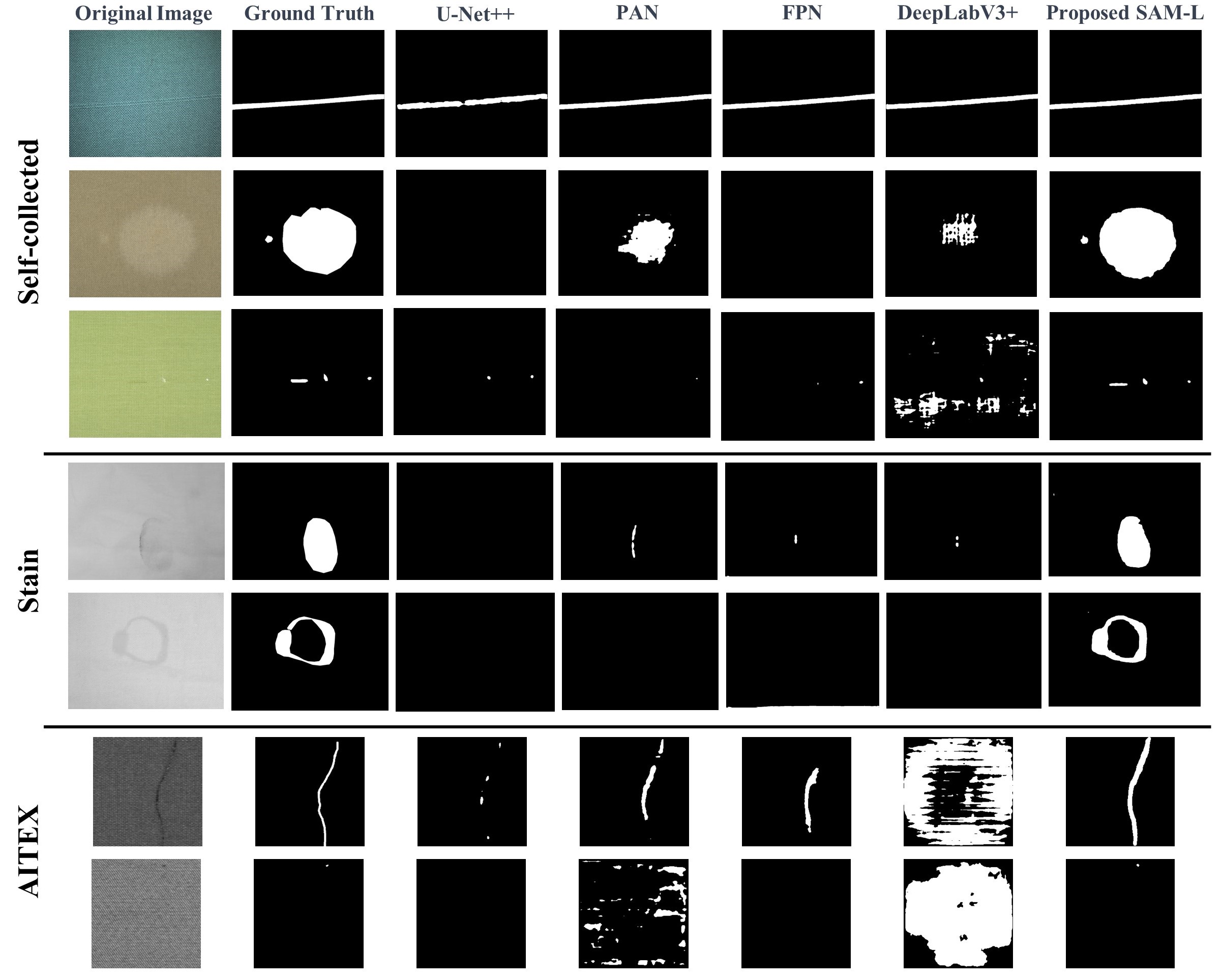}
    \caption{Comparative visualization of fabric defect segmentation results from different models. Columns from left to right respectively display: the original image, ground truth labels, predictions by U-net++, PAN, FPN, DeeplabV3+, and the proposed SAM-L model. The images showcase the superior performance of the SAM-L model in defect segmentation on three different datasets, when trained on the self-collected dataset.}
    \label{fig:result_show}
\end{figure*}

\subsection*{A. Segmentation Performance and Generalization of the Proposed SAM Model}

Figure \ref{fig:result_show} presents a range of typical test results using models trained on a self-collected training set. The self-collected results shown the predcitions on the test set, while the Stain and AITEX results are representative of cross-dataset validation performances.

In the examples from the self-collected dataset, three typical types of textile defects are analyzed. The first row shows a coarse latitude defect spanning the entire image, which is distinct and relatively easy to identify. For such conspicuous defects, both non-SAM models and our SAM model are capable of effectively recognizing and accurately delineating the defect regions. However, for more challenging defects, such as those shown in the second and third rows, our SAM model outperforms the other models significantly.

The second row shows a stain that is slightly lighter in color compared to the background, where the subtle color variation poses a challenge to defect identification. The results reveal that the non-SAM models are unable to accurately recognize the complete defect area, while our SAM model can precisely detect the entire defect, even outperforming manual annotation.

The defects shown in the third row of images include coarse latitude and two small fluffs. The complexity of detecting these defects lies in their small proportion in the image and the presence of two completely different types of defects. Non-SAM models either fail to detect the defects or can only identify one of the two types (e.g., Unet++ only detects the fluffs). In contrast, the proposed SAM model can accurately identify both types of defects within a single image.
Based on the observation of training and testing on the same dataset, it can be concluded that both non-SAM models and SAM models demonstrate good performance in recognizing easily-segmented defects. However, for challenging or less obvious defects, our proposed SAM model significantly outperforms non-SAM models.

Cross-dataset experiments primarily aim to test the model's generalization ability. First, two types of stains that result in surface color changes on the fabric surface were selected from the Stain dataset. The results show that non-SAM models are unable to effectively detect these defects, while our SAM model can accurately and completely identify the defect regions. Subsequently, we selected two types of defects from the AITEX dataset for detection. The first type is a prominent stain spanning the entire image, which non-SAM models can only partially detect while our SAM model can accurately and completely identify the entire defect. The second type of defect is more subtle, and all non-SAM models fail to identify this defect, leading to a high number of false detections, while the SAM model successfully identifies the defect without any false detections.

These examples underscore the superior generalization ability of our SAM model in two aspects. Firstly, the SAM model can effectively identify rare or unknown defects. For instance, in the self-collected dataset, there were only about five samples with Stain-type defects, and there were no defects similar to those in the Stain dataset, yet the SAM model could still precisely identify the defects in the Stain dataset. Secondly, the SAM model demonstrate excellent generalization capabilities to different backgrounds or imaging conditions. There are significant differences in images under different imaging conditions, which is why non-SAM models produced a high number of false detections on AITEX samples. However, the SAM model effectively resisted these influences and did not predict a high number of false detections on samples under different imaging conditions.

In summary, the importance of combining general knowledge and specialized knowledge to achieve excellent fabric defect segmentation performance is highlighted. Superior segmentation performance requires the model to understand and handle one or more specific types of defects, which requires the model to have a deep understanding of these defects, that's the role of specialized knowledge. However, given the diversity of types and forms of textile defects and the potential for complex situations in practical applications, relying solely on limited specialized knowledge (due to a lack of data) is insufficient. At this point, the importance of general knowledge comes into play, which includes basic skills in image processing and understanding, helping the model adapt to a changeable environment and handle a variety of defect types. In the training process of the proposed SAM model, the model not only gains specialized knowledge about textile defects through the fabric defect dataset but also inherits the general knowledge applicable to various images from the original SAM. This allows the proposed SAM model to maintain good recognition performance when facing new, unseen defects or dealing with data that is significantly different from the training dataset.

\subsection*{B. Few-shot Experiment and Its Practical Significance}

The results from our few-shot learning experiments underscore the impressive generalization capability of our proposed SAM model. As revealed by the experimental results, with 100 training images, segmentation results that are on par with those produced by the entire 909-image training set is achieved. This not only showcases the model's effectiveness but also underscores its profound practical implications.

The process of data collection and annotation often presents challenges in terms of resource and time constraints, especially in highly specialized and intricate fields such as fabric defect detection. In such instances, models that can proficiently learn from a small number of samples become particularly indispensable. The SAM model's promising performance testifies to its edge in addressing data scarcity concerns.

The fabric industry is marked by diversity, with new types of fabrics and defects frequently making their way onto the scene, necessitating models to rapidly adapt and learn. The promising performance of the SAM model under few-shot learning conditions signifies its ability to swiftly learn and accurately identify new fabric defects within a drastically shortened training cycle, thereby amplifying its effectiveness and practicality for real-world applications.

Moreover, in scenarios related to small-scale or customized production, the ability to learn from a few samples grows even more critical. In such cases, only a very limited number of samples may be available for learning. The SAM model's capacity to rapidly learn and deliver exceptional segmentation performance in such environments allows it to better cater to these real-world application scenarios.

\subsection*{C. Partial Parameter Training: Efficiency and Cost-Effectiveness}

As illustrated in the Table \ref{table:paremeter}, only a minor proportion of the total parameters are trainable, indicating the efficient utilization of computational resources.

\begin{table}[ht]
\centering
\caption{Total and Trainable Parameters of the Proposed SAM Models}
\renewcommand{\arraystretch}{1.2}
\begin{tabular}{|c|c|c|}
\hline
Model & Total Parameters & Trainable Parameters \\ 
\hline
Proposed SAM-B & 91233774 & 657442  \\
\hline
Proposed SAM-L & 309300462 & 903202  \\
\hline
Proposed SAM-H & 637515758 & 1165346  \\
\hline
\end{tabular}
\label{table:paremeter}
\end{table}

For the original SAM, training all parameters demand extensive computational resources and time. However, by focusing on just a subset of the model's parameters, the computational resources needed can be significantly reduced, thereby significantly boosting the efficiency of model training. This approach is exceptionally beneficial in research or production environments where resources are constrained.

Training only a selected subset of parameters within the fabric defect detection model yields commendable results. This strategy effectively leverages the pre-existing general knowledge embedded within the majority of pre-trained parameters. The specific adaptation to the task is handled by targeted training of a minimal subset of parameters. This methodology not only enhances training efficiency but also preserves the model's generalization ability. Consequently, the model is equipped to maintain superior performance, even when encountering previously unseen types of fabric defects.

Moreover, the tactic of training just part of the model's parameters can serve as an effective guard against overfitting. In our experiments, given the relatively limited number of training samples, overfitting is a critical concern. By limiting the number of parameters to be trained, model performance can be preserved while circumventing the loss of predictive ability for new samples due to overfitting the training data.

\subsection*{D. Annotation Correction and Assistance}

\begin{figure}[!t]
    \centering
    \includegraphics[width=3.5in]{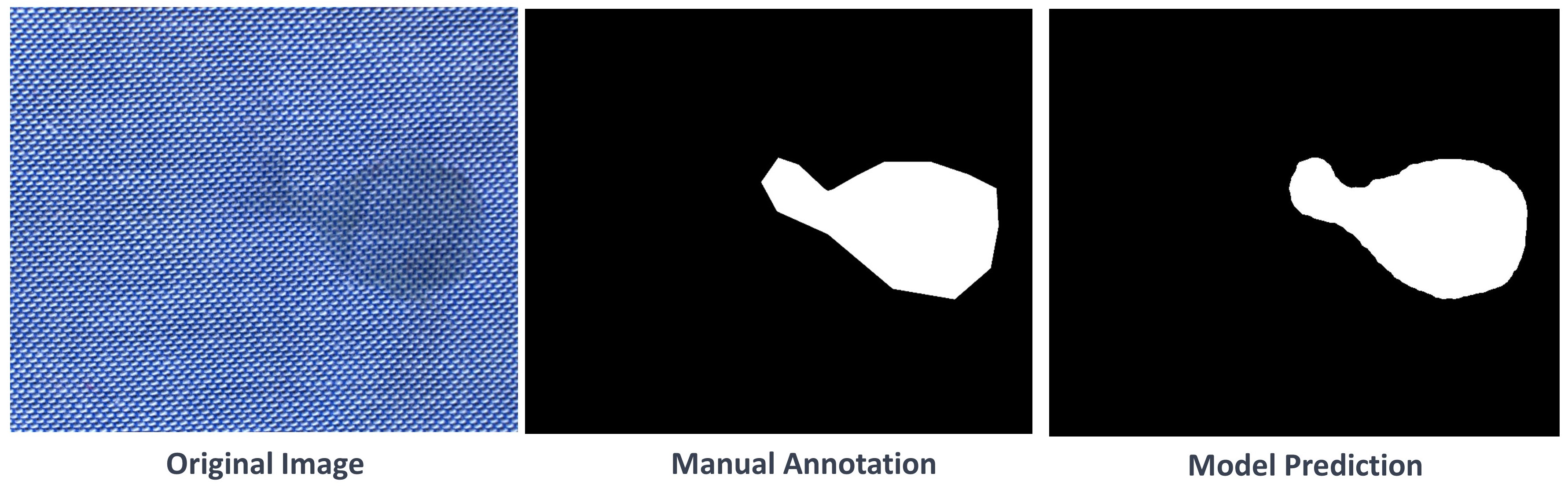}
    \caption{Comparison of Manual Annotation and Model Prediction.}
    \label{fig:compare_manual_model}
\end{figure}

Due to the robust generalization ability of the SAM model, it has a certain tolerance for errors in data annotation and will not overfit erroneous labels. As shown in Figure ~\ref{fig:compare_manual_model}, the segmentation results of SAM are superior to manual annotations, with smoother edges.

This characteristic bears significant value for practical applications. Firstly, manual annotation processes are inevitably prone to mistakes, and the error-tolerance of the SAM model effectively alleviates the impact of such mislabels on model learning. Furthermore, the SAM model can be employed to assist in manual annotation, automatically conducting preliminary defect segmentation and generating annotation suggestions, which significantly saves time and effort in manual annotation. At the same time, given that the segmentation results of the SAM model have smooth edges, for edge cases difficult to discern by the human eye, the suggestions from the SAM model can provide valuable references. Consequently, the SAM model serves not only as a tool for defect detection but also as a potent annotation tool, improving annotation efficiency, reducing annotation errors, and thereby enhancing the overall efficiency and accuracy of fabric defect detection.

\subsection*{E. Limitations and Future Direction}

This study presents an efficient way of leveraging Large Visual Models (LVMs) for fabric defect segmentation tasks. However, it is not a foundational model, as the general knowledge within the model still pertains to natural images rather than to the fabric domain. Furthermore, in this research, we have abandoned the interactive segmentation ability of the original SAM model.

For future work, we plan to explore how to inject more general knowledge of the fabric domain into the model. This may necessitate pre-training on a large-scale fabric image dataset to capture the generic knowledge of fabric textures, colors, lighting, etc. Moreover, we will investigate how to maintain the efficiency of the model while restoring the interactive segmentation capability of the SAM model, allowing users to perform fabric defect detection and annotation more flexibly.

\section{CONCLUSION}

To conclude, the present research focuses on improving the fabric defect segmentation capacity of the pre-trained SAM model. This is achieved through the integration of defect-specific knowledge via a tailored set of trainable parameters. The proposed SAM model demonstrates remarkable resilience, streamlined training, and superior performance across a range of scenarios, such as few-shot learning and cross-dataset validation. Moreover, it presents a high tolerance for annotation errors and offers assistance in annotation correction. These advancements elevate the accuracy and efficiency of fabric defect detection. This pioneering approach paves the way for considerable advancements in textile quality control, forecasting substantial resource optimization and reduction in required efforts.

\bibliographystyle{IEEEtran}
\bibliography{main}

\end{document}